\documentclass[conference]{IEEEtran}
\IEEEoverridecommandlockouts
\usepackage{marvosym}
\usepackage{cite}
\usepackage{amsmath,amssymb,amsfonts}
\usepackage{algorithmic}
\usepackage{graphicx}
\usepackage{textcomp}
\usepackage{xcolor}
\usepackage{multirow}
\usepackage{tablefootnote}
\usepackage{makecell}
\usepackage{caption} %
\def\BibTeX{{\rm B\kern-.05em{\sc i\kern-.025em b}\kern-.08em
    T\kern-.1667em\lower.7ex\hbox{E}\kern-.125emX}}
\begin{document}

\title{B\'{e}zierFormer: A Unified Architecture for 2D and 3D Lane Detection}

\author{
\IEEEauthorblockN{Zhiwei Dong, Xi Zhu, Xiya Cao, Ran Ding, Caifa Zhou, Wei Li\textsuperscript{\Letter}, Yongliang Wang, Qiangbo Liu}
\IEEEauthorblockA{Riemann lab, Huawei Technologies}
\{dongzhiwei1, caoxiya1, dingran7, zhoucaifa, liwei.levi, wangyongliang775\}@huawei.com, zhux2012@uw.edu
}
\maketitle

\begin{abstract}
Lane detection have made significant progress in recent years, but there is not a unified architecture for its two sub-tasks: 2D lane detection and 3D lane detection.
To fill this gap, we introduce B\'{e}zierFormer, a unified 2D and 3D lane detection architecture based on B\'{e}zier curve lane representation. 
B\'{e}zierFormer formulate queries as B\'{e}zier control points and incorporate a novel B\'{e}zier curve attention mechanism.
This attention mechanism enables comprehensive and accurate feature extraction for slender lane curves via sampling and fusing multiple reference points on each curve.
In addition, we propose a novel Chamfer IoU-based loss which is more suitable for the B\'{e}zier control points regression.
The state-of-the-art performance of B\'{e}zierFormer on widely-used 2D and 3D lane detection benchmarks verifies its effectiveness and suggests the worthiness of further exploration.
\end{abstract}
\section{Introduction}
\label{sec:intro}

Lane detection based on RGB images is a critical foundational perception task in various applications such as autonomous driving, lane-level AR navigation, and HD map construction. 
This task is specifically divided into 2D lane detection and 3D lane detection.
Both of them are challenging due to the inherent difficulties, including obscured or worn lane markers, and variations in lighting and weather conditions. 
With the advent of deep learning, both 2D and 3D lane detection have made significant progress \cite{xu2020curvelane,liu2021condlanenet,zheng2022clrnet,xu2022rclane,qin2022ultra,chen2022persformer,huang2023anchor3dlane,bai2023curveformer}.
However, there is not a unified architecture for these two subtasks, although they share many similarities. It usually takes a lot of efforts to adapt a state-of-the-art 2D lane detection model to 3D lane detection(or vice versa), and this hinders further development of lane detection.
\textit{In this paper, we aim to answer the question of whether it is possible to
build a unified state-of-the-art architecture for both 2D and 3D lane detection.}

{
\begin{figure}[t]
\centering
\includegraphics[width=1.0\columnwidth]{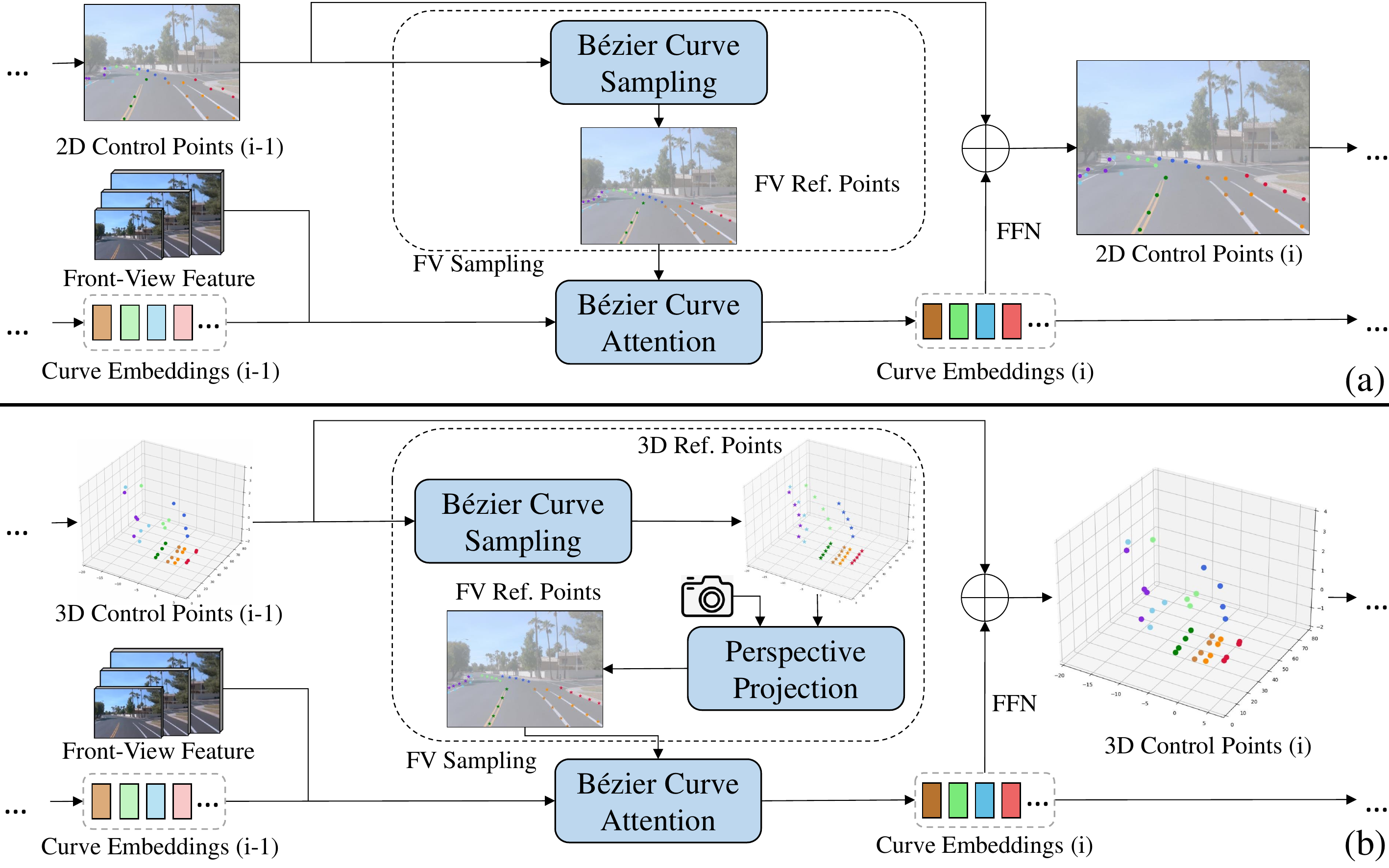}
\caption{Pipeline of B\'{e}zierFormer. We draw 2D and 3D scenarios separately for clarity. (a) B\'{e}zierFormer's decoder layer refine curve embeddings and 2D control points by extracting lane features according to input control points. (b) In 3D scenario, B\'{e}zierFormer is equipped with perspective projection.}
\label{fig_intro}
\end{figure}
}

First of all, we utilize the B\'{e}zier curve to represent lane curves in our method, because B\'{e}zier curve can uniformly and efficiently represent both 2D and 3D curves via a few control points.
Inspired by the \textbf{DE}tection \textbf{TR}ansformers \cite{carion2020end, liu2022dabdetr}, we introduce B\'{e}zierFormer, which aims to query lane features from the input monocular image and output 2D or 3D control points of lanes.
As shown in Figure \ref{fig_intro}, our method formulates queries as dynamic B\'{e}zier control points, uses a B\'{e}zier curve attention mechanism to extract lane features.
Besides we propose a Chamfer IoU-based loss compatible with 2D and 3D B\'{e}zier control points regression.


Specifically, initial control point queries, learnable curve embeddings, and image features are fed into the decoder.
With the aid of B\'{e}zier curve attention, each decoder layer extracts lane features accurately and comprehensively to refine curve embeddings and control points from the previous layer. 
To implement B\'{e}zier curve attention mechanism, the front-view (FV) sampling module first samples a few sparse reference points along the curve based on control points, ensuring that the attention's receptive field fully covers the entire lane curve. 
Then we use the B\'{e}zier curve attention operation, a multi-reference-points variant of deformable attention \cite{zhu2020deformable}, to extract and fuse the features around these reference points on the curve.
Finally, extracted lane features enhance curve embeddings for lane regression and recognition. 
For lane regression, we present a loss derived from Chamfer Distance. It facilitates a more effective learning process by fitting the overall shape of the target curve.
To make our method compatible with 2D and 3D lane detection, there is an optional perspective projection operation in the FV sampling module as depicted in Figure \ref{fig_intro}(b).
When performing 3D lane detection, the perspective projection operation uses intrinsic and extrinsic parameters of the camera to project 3D reference points, sampled based on 3D control points, onto the corresponding 2D positions on FV features for B\'{e}zier curve attention. 
Then we can reapply the whole structure in Figure \ref{fig_intro}(a) and directly detect 3D lane curves from FV features.


We conduct experiments on two widely-used 2D and 3D lane detection benchmarks. The experiments show that B\'{e}zierFormer is effective in both 2D and 3D lane detection, achieving state-of-the-art 90.72\% F1 score on CurveLanes and 58.1\% F1 score on OpenLane.

Our main contributions are summarized as follows. (1) We propose a unified 2D/3D lane detection architecture named B\'{e}zierFormer. It formulates queries as dynamic B\'{e}zier control points, introduces a novel B\'{e}zier curve attention mechanism to extract lane features accurately and comprehensively, and effectively regresses B\'{e}zier control points using a novel Chamfer IoU-based loss. (2) B\'{e}zierFormer achieves state-of-the-art performance on popular 2D and 3D lane detection benchmarks and suggests the worthiness of exploration in the future.

\section{Related Work}

\noindent \textbf{2D lane detection:} Early methods use semantic segmentation or keypoint detection to find pixels or keypoints of lane markers, then associate them to get curve instances.
These works focus on achieving better segmentation \cite{pan2018spatial,zheng2021resa} and designing better association method \cite{qu2021focus, xu2022rclane} to improve the overall lane detection performance.
These methods are intuitive but inefficient, and their detection results lack a holistic nature. 
Recently, top-down methods get more and more attention for their ability to detect lanes holistically and deal with visually challenging situations. They usually represent lanes as row-based coordinates \cite{chen2019pointlanenet,xu2020curvelane,qin2022ultra,liu2021condlanenet,zheng2022clrnet}, polynomials \cite{liu2021end} or parametric curves \cite{feng2022rethinking}, and then detect lanes in a similar way to object detection.

\noindent \textbf{3D lane detection:} Most methods transform image features from FV to bird-eye-view(BEV) for 3D lane detection. 
3D-LaneNet \cite{garnett20193d} and Gen-LaneNet \cite{guo2020gen} apply inverse perspective mapping (IPM) to transform features and utilize row-based lane anchors.
PersFormer \cite{chen2022persformer} applies deformable attention for the transformation. 
Recent works have tried to skip the feature transformation through DETR-like architecture \cite{bai2023curveformer} or 3D lane anchors \cite{huang2023anchor3dlane}.
B\'{e}zierFormer also avoids feature transformation and is a unified framework for 2D and 3D lane detection.
It should be noted that B\'{e}zierFormer is totally different with PersFormer \cite{chen2022persformer}, which just integrates distinct 2D and 3D lane detection components into a single network, instead of designing a general network for these two tasks.

\section{Method}

\begin{figure*}[htbp]
\centering
\includegraphics[width=1.0\textwidth]{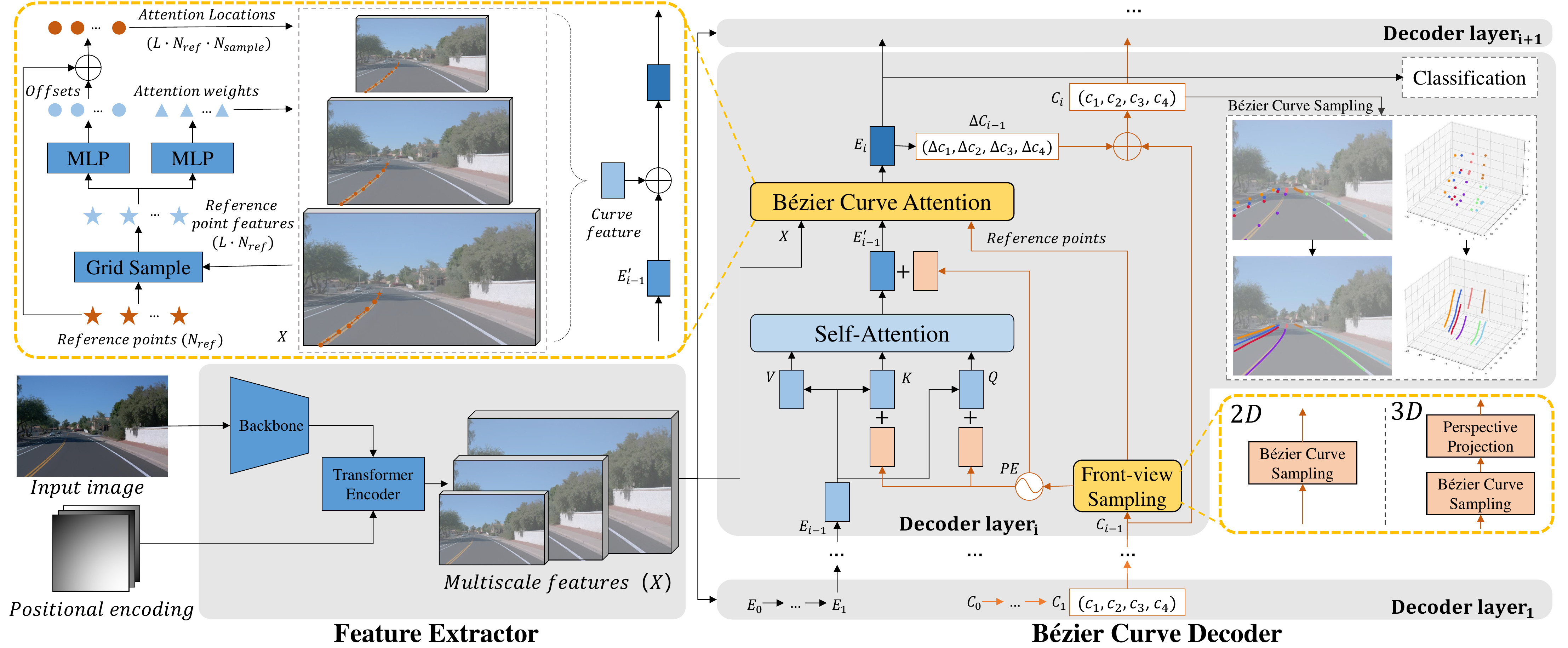} 
\caption{The architecture of B\'{e}zierFormer. The feature extractor generates multi-scale image features $X$, and the B\'{e}zier curve decoder detects lanes from $X$. Decoder layers have the same structure. Each layer receives $X$, control point queries and curve embeddings from the previous layer. The first decoder layer's input $E_{0}$ are learnable, and $C_{0}$ are generated from $X$.}
\label{fig_method}
\end{figure*}

\subsection{Lane Representation}
In B\'{e}zierFormer, we represent lane curves as B\'{e}zier Curves. 
B\'{e}zier curve of order $N$ uses $N+1$ control points $\left(c_{1}, c_{2},...,c_{N+1}\right)$  to represent a curve $S$ and is defined by:
\begin{equation}
    S(t)=\sum_{n=0}^{N}b_{n,N}(t)\cdot c_{n+1},0\le t\le1
\label{eq1}
\end{equation}
Variable $t$ ranges from 0 to 1, representing the B\'{e}zier curve being sampled from the start to the end. Coefficient $b_{n,N}(t)$ is the Bernstein basis polynomial of degree $n$ given by:
\begin{equation}
    b_{n,N}(t)=\frac{N!}{n!(N-n)!}t^{n}(1-t)^{N-n},n=0,1,...,N
\label{eq2}
\end{equation}
We adopt the classic cubic B\'{e}zier curve, which uses four control points $\left(c_{1}, c_{2}, c_{3}, c_{4}\right)$ to represent a curve.
For 2D lane detection, $c_{1},c_{2},c_{3},c_{4}$ are all 2D vectors representing the $xy$ coordinates of control points.
For 3D lane detection, they are 3D vectors representing $xyz$ coordinates.
The B\'{e}zier curve's ability to represent lane curves without relying on variables $x$, $y$ or $z$ allows for a unified representation of 2D and 3D lanes of any orientation. 

\subsection{Network Architecture}
As shown in Figure \ref{fig_method}, B\'{e}zierFormer consists of a feature extractor and B\'{e}zier curve decoder. 
The feature extractor extracts multi-scale features of input images through a backbone network and a transformer encoder containing multi-scale deformable self-attention.
The generated features are denoted as $X=\left \{x_{l}\right \}_{l=1}^{L}$, where $L$ represents the number of feature map scales.
The decoder consists of $N_{layer}$ homogeneous layers indexed starting from 1.
\textbf{Decoder layer\textsubscript{i}} receives features $X$, B\'{e}zier control point queries and corresponding curve embeddings coming from \textbf{Decoder layer\textsubscript{i-1}} as inputs. 
The B\'{e}zier control point queries model the shapes and positions of lane curves and are denoted as $C_{i-1}=\{  \left (c_{1}, c_{2}, c_{3}, c_{4}  \right )_{i-1,j} \}_{j=1}^{N_{query}}$.
$N_{query}$ is the number of queries, which means the maximum number of lane curves that B\'{e}zierFormer can detect. 
Curve embeddings contain lane features and are denoted as $E_{i-1}= \{e_{i-1,j} \}_{j=1}^{N_{query}}$.
\textbf{Decoder layer\textsubscript{i}} outputs more precise control points $C_{i}$ and better embeddings $E_{i}$ for the next layer.
For the first decoder layer, initial embeddings $E_{0}$ are randomly initialized learnable parameters. Initial control points $C_{0}$ could be initialized by $C_{0}=MLP\left(X\right)$. Because $C_{0}$ is related to the input image, this way brings a good generalization ability.

\subsection{B\'{e}zier Curve Decoder}
B\'{e}zier curve decoder consists of $N_{layer}$ layers with the same structure.
It iteratively refines curve embeddings and control point queries. 
We employ a classification head for category recognition and use B\'{e}zier curve sampling defined by Eq.(\ref{eq1}) to produce dense lane curve points for detection results.

Taking \textbf{Decoder layer\textsubscript{i}} as an example, the inputs are $X$, $C_{i-1}$, and $E_{i-1}$.
To leverage the geometric information of lane curves in self-attention and B\'{e}zier curve attention, the FV sampling module first sample $N_{ref}$ points for each query. Then, using sine positional encoding and a MLP, we fuse the positions of these $N_{ref}$ points and obtain the query's positional encoding. The formula is:
\begin{equation}
    PE(C_{i-1,j})=MLP(Concat \{PE(S_{i-1,j}(t_{k}))\}_{k=1}^{N_{ref}})
\label{eq3}
\end{equation}

After positional encoding, the relationships among lanes are modeled via  multi-head self-attention, which enhances $E_{i-1}$ to better curve embeddings $E'_{i-1}$ with global context:
\begin{equation}
    e'=e+\sum_{m=1}^{M}W_{m}\sum_{j=1}^{N_{query}}A_{m,j}W_{m}'e_{j} 
\label{eq4}
\end{equation}
$e$ and $e'$ are single embeddings in $E_{i-1}$ and $E'_{i-1}$. 
$m$ is the index of the attention head.
$W_{m}\in \mathbb{R}^{C_{e}\times C_{v}}$and $W'_{m}\in \mathbb{R}^{C_{v}\times C_{e}}$ are learnable parameters, where $C_{e}$ and $C_{v}=C_{e}/M$ are the feature dimensions of the embedding and the key in attention respectively.
The attention weights $A_{m,j} \propto \exp\{\frac{e_{j}^{T}U_{m}^{T}V_{m}e_{j}}{\sqrt{C_{v}}}\}$ are normalized to $\sum_{j=1}^{N_{query}}A_{m,j}=1$, and $U_{m},V_{m}\in \mathbb{R}^{C_{v}\times C_{e}}$ are learnable parameters.

Subsequent B\'{e}zier curve attention employs the $N_{ref}$ points produced by FV sampling module as reference points.
It collects curve features of each query from $X$ and updates $E'_{i-1}$ to better embeddings $E_{i}$ for the next layer. 
A classification head calculates the category vectors $V_{i}$ from $E_{i}$.
A simple FFN predicts the offset $\Delta C_{i}$, then the control points are refined as $C_{i}=\Delta C_{i}+C_{i-1}$.
To achieve layer-by-layer refinement, $V_{i}$ and $C_{i}$ of each layer need to be supervised.

\subsection{B\'{e}zier Curve Attention Mechanism}
B\'{e}zier curve attention mechanism aims to better capture features of the slender lane curves represented by B\'{e}zier curves.
We refrain from directly using vanilla deformable attention due to two primary reasons that render it suboptimal for extracting slender lane features.
Firstly, deformable attention originates from object detection and generates only one reference point per object, but one reference point is insufficient to describe a slender lane curve comprehensively. 
Secondly, the reference points of deformable attention are adaptively generated, which causes slow convergence and less accurate lane feature extraction.
To overcome these shortcomings, B\'{e}zier curve attention mechanism leverages B\'{e}zier curve sampling and B\'{e}zier curve attention operation to sample and fuse multiple reference points along the curve for accurate and comprehensive lane features.

The B\'{e}zier curve attention operation does not involve any interaction among $\{ {e'}_{i-1,j}\}_{j=1}^{N_{query}}$, so we simplify the explanation of this operation by considering the update of a single $e'_{i-1,j}$.
Let $\{ r_{k} \}_{k=1}^{N_{ref}}$ be the $N_{ref}$ reference points. As shown in the left-top part of Figure \ref{fig_method}, the features $\{ \{ f_{k,l} \}_{k=1}^{N_{ref}} \}_{l=1}^{L}$ of the reference points at different scales of $X$ are sampled, then two MLPs calculate $f_{k,l}$ to obtain $N_{sample}$ attention location offsets $\{ o_{k,l,s} \}_{s=1}^{N_{sample}}$ and attention weights $\{ w_{k,l,s} \}_{s=1}^{N_{sample}}$ for $r_{k}$ at scale $l$. 
$\{ o_{k,l,s} \}_{s=1}^{N_{sample}}$ are added to $r_{k}$ to get the attention locations $\{ p_{k,l,s} \}_{s=1}^{N_{sample}}$. To update $e'_{i-1,j}$, the formula is:
\begin{equation}
    \begin{split}
        e=e'+\sum_{m=1}^{M}W_{m}\sum_{k=1}^{N_{ref}}\sum_{l=1}^{L}\sum_{s=1}^{N_{sample}}W_{m,k,l,s}W'_{m}x_{l}(p_{k,l,s})
    \end{split}
\label{eq5}
\end{equation}
For simplicity, $e'$ and $e$ represent $e'_{i-1,j}$ and $e_{i,j}$, respectively. 
$m$, $W_{m}$ and $W'_{m}$ have the same meanings as they have in Eq.(\ref{eq4}).
The dynamically generated attention weights $W_{m,k,l,s}$ are normalized to $\sum_{k=1}^{N_{ref}}\sum_{s=1}^{N_{sample}}W_{m,k,l,s}=1$.
B\'{e}zier curve attention operation extracts lane curve features from FV features and updates curve embeddings $E'_{i-1}$ via Eq.(\ref{eq5}). Enhanced curve embeddings $E_{i}$ are subsequently used for control points refinement and category recognition.

\subsection{Loss Function and Label Assignment}
We present a lane regression loss derived from Chamfer Distance (CD) to better regress B\'{e}zier curve and use Focal Loss \cite{lin2017focal} for classification. 
\cite{feng2022rethinking} shows that regressing the sampled points can achieve better results than directly regressing B\'{e}zier control points. 
However, this method may sometimes be limited to optimizing the distance between local points, while ignoring the overall shape and location fit between two curves.
We give an intuitive explanation of this limitation in the Appendix.
We propose calculating CD between curves, as it considers each curve a whole entity to get the shortest distance from a point to the curve.
As CD is calculated between point sets instead of continuous curves, $N_{dis}$ points $ \{ p_{i}^{A} \}_{i=1}^{N_{dis}}$ and $ \{ p_{i}^{B} \}_{i=1}^{N_{dis}}$ on curve $A$ and $B$ are sampled.
Inspired by \cite{zheng2022clrnet}, we give the lane curve a width $e$ and normalize CD to IoU, which we call Chamfer IoU, or CIoU for short. It is defined by:
\begin{equation}
    CIoU_{A\to B}=\frac{1}{N_{dis}}\sum_{i=1}^{N_{dis}} \frac{2e-min\left \{ L2\left (p_{i}^{A}, p_{j}^{B} \right )  \right \} _{j=1}^{N_{dis}} }{2e+min\left \{ L2\left (p_{i}^{A}, p_{j}^{B} \right )  \right \} _{j=1}^{N_{dis}} } 
\label{eq6}
\end{equation}
The regression loss between a pair of predicted curve $S_{pred}$ and ground truth curve $S_{gt}$ can be formulated as:
\begin{equation}
    L_{loc}=1-\frac{1}{2} \left ( CIoU_{S_{pred}\to S_{gt}} + CIoU_{S_{gt}\to S_{pred}}\right ) 
\label{eq7}
\end{equation}
However, the computation of CIoU ignores the order among the sampled points. To ensure the correct order of the sampled points, we add two simple geometric constraints.
\begin{equation}
    L_{len}=L1(\frac{len(S_{pred})}{len(S_{gt})},1) 
\label{eq8}
\end{equation}
\begin{equation}
    L_{endpoint}=\frac{1}{2} \left [L2(\hat{p}_{1}^{S_{pred}},\hat{p}_{1}^{S_{gt}})+L2(\hat{p}_{N_{dis}}^{S_{pred}},\hat{p}_{N_{dis}}^{S_{gt}})\right ] 
\label{eq9}
\end{equation}
$len()$ means calculating the curve length. $\hat{p}_{1}, \hat{p}_{N_{dis}}$ represent the normalized endpoints of the curve. Then, we get regression loss as $L_{reg}=L_{loc}+L_{len}+L_{endpoint}$
and total loss is the sum of the losses of all decoder layers. The formula is:
\begin{equation}
    L_{total}=\left ( L_{reg}^{0}+L_{cls}^{0}\right ) + \sum_{i=1}^{N_{layer}}\left ( L_{reg}^{i}+L_{cls}^{i} \right ) 
\label{eq11}
\end{equation}
To match predictions with ground truth, we adopt SimOTA \cite{ge2021yolox} to dynamically assign $topk$ predictions to each ground truth with the matching cost $cost_{i,j}=L_{reg_{i,j}}+L_{cls_{i,j}}$.
Specifically, we set $topk=1$ for an NMS-free pipeline.


\section{Experiments}

\noindent \textbf{Datasets:} We conduct experiments on two widely-used 2D and 3D lane detection benchmarks.
\textbf{CurveLanes} \cite{xu2020curvelane} is a large-scale 2D lane dataset with complex lane topologies. 
It comprises 150,000 images for training, validation, and testing.
In CurveLanes, curved lanes account for over 90\%, encompassing challenging cases such as dense, forked, merged, and nearly horizontal lanes.
\textbf{OpenLane} \cite{chen2022persformer} is a recently released large-scale real-world 3D lane dataset, with a total of 200,000 images and over 880,000 3D annotations.

\noindent \textbf{Evaluation Metric:} To keep consistent with \cite{xu2020curvelane}, we use F1 score for 2D lane detection evaluation. 
For OpenLane, we follow \cite{chen2022persformer} to report the F1 score and category accuracy.

\noindent \textbf{Implementation Details:} We choose ResNet18 \cite{He_2016_CVPR} and Swin-Tiny \cite{Liu_2021_ICCV} as pre-trained backbones.
Input images are resized to $800\times320$ on CurveLanes, and $480\times360$ on OpenLane.
We set $N_{query}$ for CurveLanes and OpenLane to 16 and 32, respectively. 
We train 24 epochs with batch size 16, using an AdamW optimizer with a learning rate of 1e-4. 
We set $t$ to 0, 0.25, 0.5, 0.75, and 1 to obtain $N_{ref}=5$ reference points, and $N_{sample}=5$.
To calculate $L_{reg}$, we set $N_{dis}=200$ and lane width $e=10$ for CurveLanes, $e=0.9$ for OpenLane.
More details are in the Appendix.

\subsection{Comparisons with the State-of-the-Art Methods}


\subsubsection{2D lane detection}

Table \ref{tab_curvelane} shows B\'{e}zierFormer achieves top-tier performance on CurveLanes.
Even based on ResNet18, B\'{e}zierFormer achieves a better F1 score of 89.06\% than CLRNet's 86.48\% based on ResNet101.
With Swin-Tiny as the backbone, B\'{e}zierFormer achieves a state-of-the-art F1 score of 91.06\%, which is 4.58\% higher than that of CLRNet.
Compared to the best bottom-up method RCLane, B\'{e}zierFormer achieves a comparable performance but a much higher inference efficiency.
The results on CurveLanes indicate that B\'{e}zierFormer is effective and effecient in 2D lane detection. 
We give more qualitative visualized results in the Appendix.

\begin{table}[htbp]
  \centering
    \small
    \setlength{\tabcolsep}{4pt}
    \begin{tabular}{ll|ccccc}
    \Xhline{1px}
    \multicolumn{1}{l}{Method} & \multicolumn{1}{l|}{Backbone} & \ \ F1 (\%)    & P (\%) & R (\%) & FPS\\
    \Xhline{0.5px}
    SCNN\cite{pan2018spatial}\textsuperscript{\dag}  & VGG16  & 65.02 & 76.13 & 56.74 & 8\\
    PointLaneNet\cite{chen2019pointlanenet}\textsuperscript{\dag} & ResNet101  & 78.47 & 86.33 & 72.91 & -\\
    CurveLane-L\cite{xu2020curvelane}\textsuperscript{\dag} & \ \ \ \ \ \ \ -  & 82.29 & 81.11 & 75.03 & -\\
    CondLaneNet\cite{liu2021condlanenet}\textsuperscript{\ddag} & ResNet101  & 86.1  & 88.98 & 83.41 & 48\\
    UFLDv2\cite{qin2022ultra}\textsuperscript{\ddag} & ResNet34  & 81.34  & 81.93 & 80.76 & 86\\
    CLRNet\cite{zheng2022clrnet}\textsuperscript{\S} & ResNet101  & 86.48 & 91.77 & 81.76 & 74\\
    RCLane\cite{xu2022rclane}\textsuperscript{\ddag}   &SegFormer   & \textbf{91.43} & \textbf{93.96}  & \textbf{89.03} & 25\\
    \Xhline{0.5px}
    B\'{e}zierFormer & ResNet18  & 89.06 & 93.57 & 84.96 & \textbf{133}\\
    B\'{e}zierFormer & Swin-Tiny  &
    \textbf{91.31} & \textbf{93.81} & \textbf{88.94} & 84\\
    \Xhline{1px}
    \end{tabular}%
     \caption{Results on CurveLanes. P and R mean precision and recall. \textsuperscript{\dag} means the results are from \cite{xu2020curvelane}. \textsuperscript{\ddag} means the results are from the corresponding original papers, and \textsuperscript{\S} means the results are reproduced based on official code.}
  \label{tab_curvelane}%
\end{table}%

\subsubsection{3D lane detection}
Table \ref{tab_openlane} shows that B\'{e}zierFormer achieves state-of-the-art performance on OpenLane. 
With the input resolution of $480\times360$, B\'{e}zierFormer achieves an F1 score of 56.43\% and a category accuracy of 94.1\%, which are 2.73\% and 3.2\% higher than those of the second-best Anchor3DLane, and has a higher FPS.
It is worth noting that NMS-free B\'{e}zierFormer only uses 32 queries, while Anchor3Dlane needs over 4,000 3D lane anchors. 
CurveFormer is also a DETR-like method but represents lane curves as 3D point sets. 
B\'{e}zierFormer outperforms CurveFormer by 5.93\% F1 score and is superior in all scenarios and X/Z errors, indicating that B\'{e}zier curve is a better representation of 3D lane curves than 3D point sets. 
With a larger input resolution of $1024\times576$, B\'{e}zierFormer\textsuperscript{$\star$} achieves a higher F1 score of 58.6\% and a category accuracy of 94.2\%. 
Tabele \ref{tab_openlane} demonstrates B\'{e}zierFormer's effectiveness and efficiency in 3D lane detection.
More qualitative results of B\'{e}zierFormer are in the Appendix.

\begin{table}[htbp]
  \centering
    \small
    \setlength{\tabcolsep}{4pt}
    \begin{tabular}{ll|cc|c}
    \Xhline{1px}
    \multicolumn{1}{l}{Method} & \multicolumn{1}{l|}{Backbone} & F1(\%) & Cate Acc(\%) & FPS \\
    \Xhline{0.5px}
    3D-LaneNet\cite{garnett20193d}\textsuperscript{\dag} & VGG16 & \centering 44.1  & \centering -  & -\\
    Gen-LaneNet\cite{guo2020gen}\textsuperscript{\dag} & ERFNet & \centering 32.3  & \centering -     & -\\
    PersFormer\cite{chen2022persformer}\textsuperscript{\ddag} & EfficientNet & \centering 50.5  & \centering 92.3  & 15 \\
    CurveFormer\cite{bai2023curveformer}\textsuperscript{\ddag} & EfficientNet & \centering 50.5  & \centering -     & -\\
    Anchor3DLane\cite{huang2023anchor3dlane}\textsuperscript{\ddag} & ResNet18 & \centering 53.7  & \centering 90.9  & 74\\
    \Xhline{0.5px}
    B\'{e}zierFormer & ResNet18 & \centering 53.84 & \centering 92.02 & \textbf{151}\\
    B\'{e}zierFormer & Swin-Tiny & \centering 56.43 & \centering 94.1  & 90\\
    B\'{e}zierFormer\textsuperscript{$\star$} & Swin-Tiny & \centering \textbf{58.6} & \centering \textbf{94.2} & 42\\
    \Xhline{1px}
    \end{tabular}%
    \captionsetup{justification=raggedright,singlelinecheck=false}
    \caption{Results on OpenLane. \textsuperscript{$\star$} means the input resolution is $1024\times576$, and other methods use $480\times360$. \textsuperscript{\dag} means the results are from \cite{chen2022persformer}. \textsuperscript{\ddag} has the same meaning as in Table \ref{tab_curvelane}.}
  \label{tab_openlane}%
\end{table}%


\subsection{Ablation Studies}
In this section, we conduct all the studies with ResNet18.

\subsubsection{Lane Representation}

To demonstrate the superiority of B\'{e}zier curve in lane detection, we compare it with polynomial and row-based representations.
The polynomial and row-based representations are referenced to LSTR\cite{liu2022learning} and CurveFormer\cite{bai2023curveformer}, respectively. 
For fairness, we adapt these representations into an architecture akin to B\'{e}zierFormer, detailed in the Appendix, and use advanced LineIoU loss \cite{zheng2022clrnet} for better performance. 
Table \ref{tab_rep} reveals that B\'{e}zier curve yields much better results. 



\begin{table}[htbp]
    \centering
    \setlength{\tabcolsep}{4pt}
    \begin{tabular}{l|ccc}
    \Xhline{1px}
    Method & Poly & Row-based & B\'{e}zier Curve \\
    \Xhline{0.5px}
    $F1_{CurveLanes}$ (\%)  & 80.90 & 83.10 & \textbf{89.06}    \\ 
    $F1_{OpenLane}$ (\%)  & 50.30 & 51.27 & \textbf{53.84}   \\ 
    \Xhline{1px}
    \end{tabular}
\caption{Comparison of different lane representations.}
\label{tab_rep}
\end{table}

\subsubsection{Attention mechanism}
\begin{figure}[hbtp]
\centering
\includegraphics[width=0.9\columnwidth]{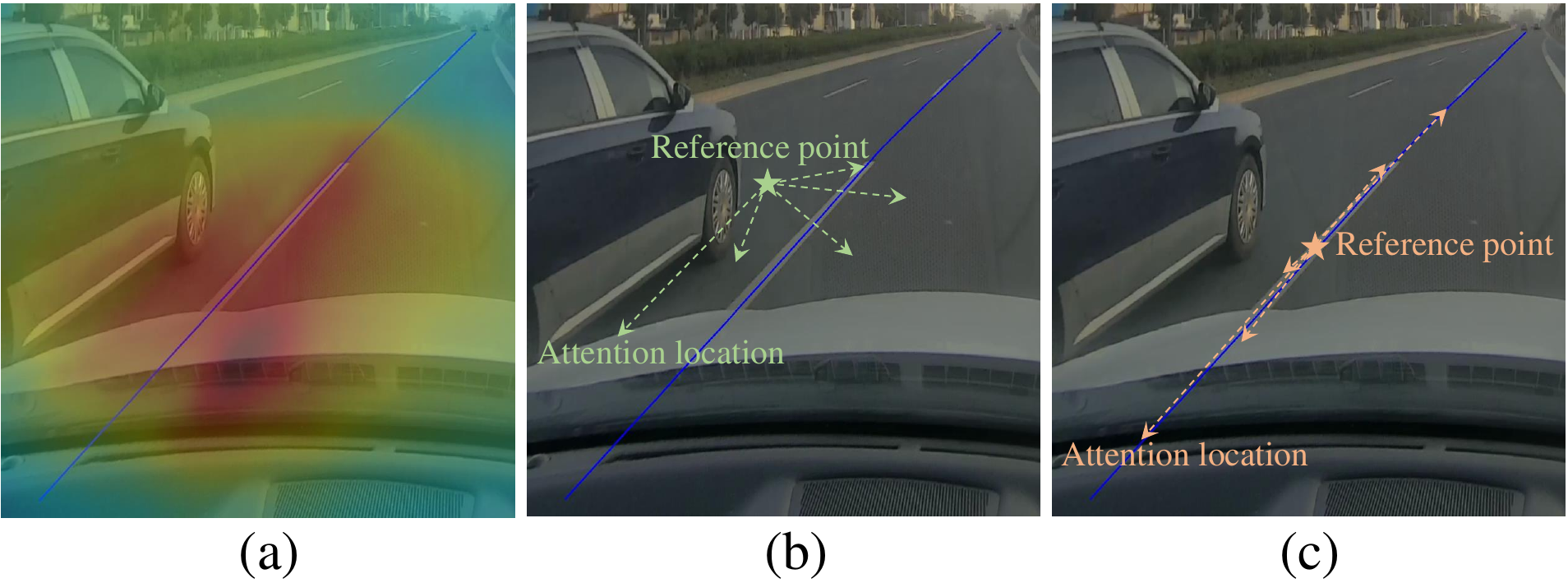}
\caption{Attention visualization. (a) Ordinary attention. (b) Deformable attention. (c) B\'{e}zier curve attention.}
\label{fig_att}
\end{figure}

To validate the superiority of our B\'{e}zier curve attention mechanism in lane detection, we compare it with ordinary attention \cite{vaswani2017attention} and vanilla deformable attention.
Vanilla deformable attention only adaptively generates one reference point, so B\'{e}zier curve attention uses $t=0.5$ to sample one reference point for fairness.
As shown in Table \ref{tab_att}, B\'{e}zier curve attention mechanism obviously brings better performance through explicitly sampling reference points.
Figure \ref{fig_att} also illustrates that B\'{e}zier curve attention has more precise attention locations than the other two attention mechanisms.

\begin{table}[htbp]
  \centering
  \setlength{\tabcolsep}{2pt}
    \begin{tabular}{l|ccc}
    \Xhline{1px}
    Method & Ordinary & Deformable & B\'{e}zier Curve \\
    \Xhline{0.5px}
    $F1_{CurveLanes}$(\%)    & 82.76 & 85.81 & \textbf{87.60} \\
    $F1_{OpenLane}$(\%)    & 46.53 & 49.08 & \textbf{52.13} \\
    \Xhline{1px}
    \end{tabular}%
    \caption{Comparison of different attention mechanisms.}
  \label{tab_att}%
\end{table}%

\subsubsection{Loss Function}
On CurveLanes, we compare our Chamfer IoU-based regression loss with the sampling loss \cite{feng2022rethinking} for B\'{e}zier curve fitting. 
For fairness, we normalize the sampling distance between curves' points to IoU and get sampling IoU loss. 
Table \ref{tab_loss} indicates that our regression loss performs better because Chamfer IoU measures overall shape and location similarity between curves better.

\begin{table}[htbp]
  \centering
  \setlength{\tabcolsep}{4pt}
    \begin{tabular}{l|ccc}
    \Xhline{1px}
    Method & Sampling & Sampling IoU & Ours \\
    \Xhline{0.5px}
    $F1_{CurveLanes}$(\%)    & 78.78 & 81.86 & \textbf{89.06} \\
    $F1_{OpenLane}$(\%)    & 47.11 & 48.32 & \textbf{53.84} \\
    \Xhline{1px}
    \end{tabular}%
    \caption{Comparison of different lane regression losses.}
  \label{tab_loss}%
\end{table}%

\section{Conclusion}
In this work, we introduce B\'{e}zierFormer, a novel unified 2D/3D lane detection architecture. 
It employs B\'{e}zier control point queries to represent lane curves and uses a novel B\'{e}zier curve attention mechanism to accurately and comprehensively extract lane features.
Besides, we propose the Chamfer IoU-based regression loss to improve the performance.
Finally, its state-of-the-art performance on widely-used 2D and 3D lane detection benchmarks attests to B\'{e}zierFormer's effectiveness in both 2D and 3D lane detection and suggests that further exploration would be valuable.


\section{Appendix}
\subsection{Implementation Details}
We implement B\'{e}zierFormer based on MMDetection \cite{mmdetection}.
For reproducibility, we give the detailed experimental settings, including hyperparameters of the training and testing process, data augmentation strategies, and network settings of B\'{e}zierFormer in Table \ref{tab_exp}.

\begin{table*}[]
  \centering
    \begin{tabular}{l|p{10.065em}p{10.065em}p{10.065em}}
    \hline
    Dataset & \multicolumn{1}{c}{CULane} & \multicolumn{1}{c}{CurveLanes} & \multicolumn{1}{c}{OpenLane} \\
    \hline
    Input resolution & \multicolumn{1}{c}{$800 \times 320$} & \multicolumn{1}{c}{$800 \times 320$} & \multicolumn{1}{c}{$1024 \times 576$} \\
    Epochs & \multicolumn{1}{c}{24} & \multicolumn{1}{c}{24} & \multicolumn{1}{c}{24} \\
    Batch size & \multicolumn{1}{c}{16} & \multicolumn{1}{c}{16} & \multicolumn{1}{c}{16} \\
    Optimizer & \multicolumn{1}{c}{AdamW} & \multicolumn{1}{c}{AdamW} & \multicolumn{1}{c}{AdamW} \\
    LR    & \multicolumn{1}{c}{0.0001} & \multicolumn{1}{c}{0.0001} & \multicolumn{1}{c}{0.0001} \\
    LR of backbone & \multicolumn{1}{c}{0.00001} & \multicolumn{1}{c}{0.00001} & \multicolumn{1}{c}{0.00001} \\
    LR decay & \multicolumn{1}{c}{Poly} & \multicolumn{1}{c}{Poly} & \multicolumn{1}{c}{Poly} \\
    LR decay setting & \multicolumn{1}{c}{power=2,min=1e-5} & \multicolumn{1}{c}{power=2,min=1e-5} & \multicolumn{1}{c}{power=2,min=1e-5} \\
    Weight decay & \multicolumn{1}{c}{0.0001} & \multicolumn{1}{c}{0.0001} & \multicolumn{1}{c}{0.0001} \\
    Warmup epochs & \multicolumn{1}{c}{None} & \multicolumn{1}{c}{None} & \multicolumn{1}{c}{None} \\
    \hline
    Horizontal flip & \multicolumn{1}{c}{yes} & \multicolumn{1}{c}{yes} & \multicolumn{1}{c}{no} \\
    RandomAffine & \multicolumn{1}{c}{yes} & \multicolumn{1}{c}{yes} & \multicolumn{1}{c}{no} \\
    Color jitter & \multicolumn{1}{c}{yes} & \multicolumn{1}{c}{yes} & \multicolumn{1}{c}{no} \\
    Blur  & \multicolumn{1}{c}{yes} & \multicolumn{1}{c}{yes} & \multicolumn{1}{c}{no} \\
    RandomBrightness & \multicolumn{1}{c}{yes} & \multicolumn{1}{c}{yes} & \multicolumn{1}{c}{no} \\
    \hline
    Control points dimension & \multicolumn{1}{c}{2} & \multicolumn{1}{c}{2} & \multicolumn{1}{c}{3} \\
    Perspective projection & \multicolumn{1}{c}{no} & \multicolumn{1}{c}{no} & \multicolumn{1}{c}{yes} \\
    N\textsubscript{query} & \multicolumn{1}{c}{10} & \multicolumn{1}{c}{16} & \multicolumn{1}{c}{32} \\
    N\textsubscript{ref} & \multicolumn{1}{c}{5} & \multicolumn{1}{c}{5} & \multicolumn{1}{c}{5} \\
    N\textsubscript{sample} & \multicolumn{1}{c}{5} & \multicolumn{1}{c}{5} & \multicolumn{1}{c}{5} \\
    N\textsubscript{dis} & \multicolumn{1}{c}{200} & \multicolumn{1}{c}{200} & \multicolumn{1}{c}{200} \\
    e(width of lane curve) & \multicolumn{1}{c}{10} & \multicolumn{1}{c}{10} & \multicolumn{1}{c}{0.9} \\
    Dimension of E\textsubscript{i} & \centering {Res18:128,Swin-T:256} & \multicolumn{1}{p{10.75em}}{Res18:128,Swin-T:256} & \multicolumn{1}{p{9.875em}}{Res18:256,Swin-T:256} \\
    Number of feature scales & \multicolumn{1}{c}{4} & \multicolumn{1}{c}{4} & \multicolumn{1}{c}{4} \\
    Number of decoder layers & \multicolumn{1}{c}{6} & \multicolumn{1}{c}{6} & \multicolumn{1}{c}{6} \\
    \hline
    \end{tabular}%
    \caption{Detailed experiment settings of B\'{e}zierFormer.}
  \label{tab_exp}%
\end{table*}%

\subsection{Illustration of Loss Function}

\begin{figure}[htbp]
\centering
\includegraphics[width=0.95\columnwidth]{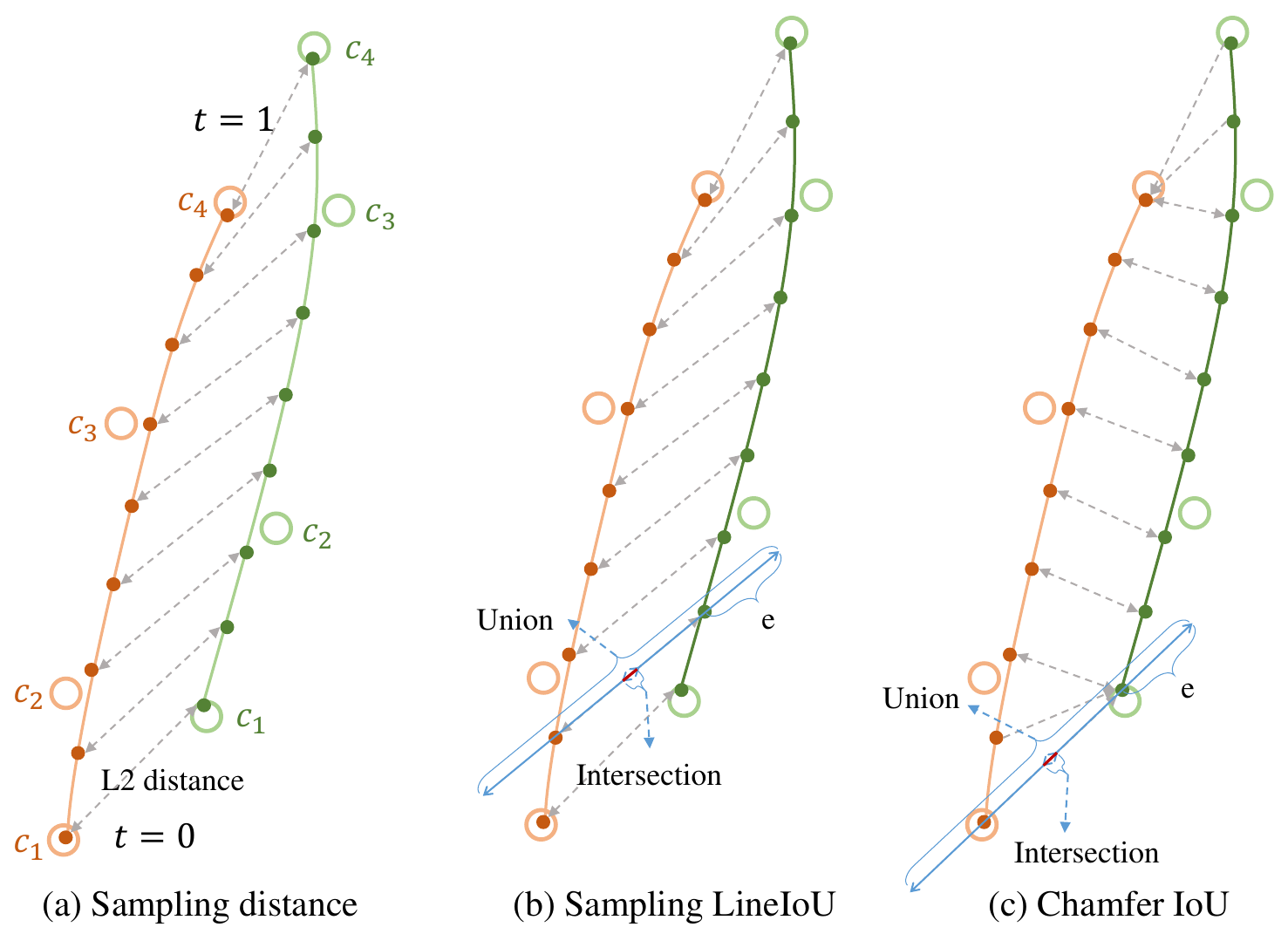} 
\caption{Illustration of different curve regression losses.}
\label{fig_loss}
\end{figure}

For a more intuitive understanding of the advantage of Chamfer IoU-based lane regression loss, Figure \ref{fig_loss} visualizes different losses. 
Figure \ref{fig_loss}(c) shows that the distances between two curves' endpoints are relatively larger than other locations, contributing to a more effective learning process compared with Figure \ref{fig_loss}(a) and Figure \ref{fig_loss}(b).

\subsection{Poly and Row-based Curve Decoder}
\begin{figure*}[h]
\centering
\includegraphics[width=1.0\textwidth]{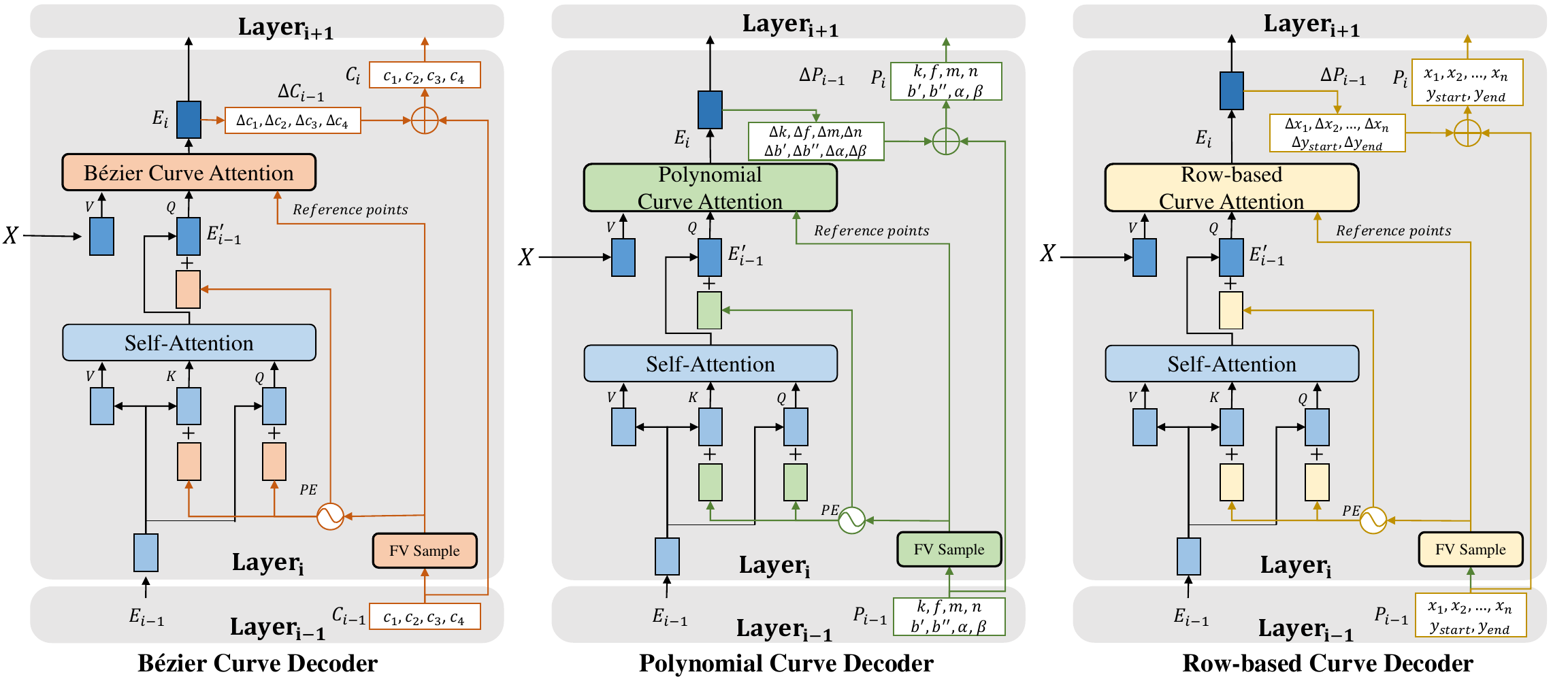} 
\caption{The architecture of three different decoders.}
\label{fig_dec}
\end{figure*}

For a fair comparison with polynomial and row-based lane representation, we employ the same feature extractor and design delicated decoders for them, similar with B\'{e}zier curve decoder as shown in Figure \ref{fig_dec}. 

For polynomial representation, we refer to LSTR \cite{liu2021end} and use $\left(k,f,m,n,b^{'},b^{''},\alpha ,\beta \right)$ to formulate a curve. LSTR uses vertical $y$ as variable to describe a curve, $k,f,m,n,b^{'},b^{''}$ is the polynomial parameters, $\alpha ,\beta$ represents the $y$ coordinates of endpoints. The formula is:
\begin{equation}
    x=\frac{k}{\left(y-f \right)^{2}} + \frac{m}{\left(y-f \right)} + b^{'}\times y-b^{''}
    \label{eq_1}
\end{equation}
To sample the reference points for \textbf{Poly Curve Attention}, which has the same equation with \textbf{B\'{e}zier Curve Attention}, we selects five equally-spaced $y$ coordinates:
\begin{equation}
    y_{i}=\alpha + \frac{\beta - \alpha}{4}\cdot i, i=0,1,2,3,4
    \label{eq_2}
\end{equation}
Thus, we can get the reference points of \textbf{Poly Curve Attention} according to Eq.(\ref{eq_1}). Finally, the reference points are:

\begin{equation}
    \left ( x_{i},y_{i} \right ), i=0,1,2,3,4
\end{equation}

For row-based representation, we refer to Laneformer \cite{han2022laneformer}. Laneformer formulates a lane curve as $\left (x_{1},x_{2},...,x_{72},y_{start},y_{end} \right )$, where $\left (x_{1},x_{2},...,x_{72}\right ) $ are the $x$ coordinates for the 72 equally-spaced $y$ coordinates, and $y_{start},y_{end}$ denote for the start $y$ coordinate and end $y$ coordinate of the curve. \textbf{Row-based Curve Attention} also shares the same equation with \textbf{B\'{e}zier Curve Attention}, but has a different reference points sampling method. To get the five reference points, we compute five equally-spaced indexes of $x$ according to $y_{start}$ and $y_{end}$, as follows.

\begin{equation}
    idx_{i}=\left \lfloor y_{start} + \frac{y_{end}-y_{start}}{4} \cdot i  \right \rfloor, i=0,1,2,3,4
\end{equation}

Then we can get the five reference points:
\begin{equation}
    \left ( x_{idx_{i}},idx_{i} \right ), i=0,1,2,3,4
\end{equation}

\subsection{Results on CULane}
CULane \cite{pan2018spatial} is also a famous large-scale 2D lane detection dataset including nine challenging scenarios on highways and urban roads. CULane's challenge lies in various difficult visual scenes. Although B\'{e}zierFormer is not designed with this challenge in mind, it still slightly outperforms methods which are proposed to address this challenge, such as CLRNet \cite{zheng2022clrnet}, on CULane. 

\begin{table*}[htbp]
  \centering
    \small
    \setlength{\tabcolsep}{2.5pt}
    \begin{tabular}{ll|ccc|ccccccccc}
    \Xhline{1px}
    \multicolumn{1}{l}{Method} & \multicolumn{1}{l|}{Backbone}  &  mF1(\%)   & F1(\%) & F1\textsubscript{@75}(\%) & Normal & Crowd & Hlight & Shadow & NoLine & Arrow & Curve &  Cross$\downarrow$ & Night \\
    \Xhline{0.5px}
    \textbf{Bottom-Up:} &       &       &       &       &       &       &       &       &       &       &       &       &       \\
    RESA\cite{zheng2021resa}\textsuperscript{\dag}  & ResNet50 &  47.86 & 75.3  & 53.39 & 92.1  & 73.1  & 69.2  & 72.8  & 47.7  & 88.3  & 70.3  & 1503  & 69.9  \\
    FOLOLane\cite{qu2021focus}\textsuperscript{\ddag} & ERFNet &  -     & 78.8  & -     & 92.7  & 77.8  & 75.2  & 79.3  & 52.1  & 89    & 69.4  & 1569  & 74.5 \\
    GANet\cite{wang2022keypoint}\textsuperscript{\S} & ResNet101 &  54.71 & 79.63 & 62.33 & 93.67 & 78.66 & 71.82 & 78.32 & 53.38 & 89.86 & 77.37 & 1352  & 73.85 \\
    RCLane\cite{xu2022rclane}\textsuperscript{\ddag} & SegFormer &  -     & \textbf{80.5}  & -     & 94.01 & 79.13 & 72.92 & 81.16 & 53.94 & 90.51 & 79.66 & 931  & 75.1 \\
    \Xhline{0.5px}
    \textbf{Top-Down:} &       &  &       &       &       &       &       &       &       &       &       &       &       \\
    UFLDv2\cite{qin2022ultra}\textsuperscript{\S} & ResNet34 &  49.94 & 76    & 55.49 & 92.5  & 74.8  & 65.5  & 75.5  & 49.2  & 88.8  & 70.1  & 1910  & 70.8  \\
    LaneATT\cite{tabelini2021keep}\textsuperscript{\dag} & ResNet122 &  51.48 & 77.02 & 57.5  & 91.74 & 76.16 & 69.47 & 76.31 & 50.46 & 86.29 & 64.05 & 1264  & 70.81 \\
    LSTR\cite{liu2022learning}\textsuperscript{\S}  & ResNet18 &  35.93 & 68.72 & 34.23 & 86.78 & 67.34 & 56.63 & 59.82 & 40.1  & 78.66 & 56.64 & 1166  & 59.92 \\
    CondLaneNet\cite{liu2021condlanenet}\textsuperscript{\dag} & ResNet101 &  54.83 & 79.48 & 61.23 & 93.47 & 77.44 & 70.93 & 90.91 & 54.13 & 90.16 & 75.21 & 1201  & 74.8  \\
    LaneFormer\cite{han2022laneformer}\textsuperscript{\ddag} & ResNet50 &  -     & 77.06 & -     & 91.77 & 75.41 & 70.17 & 75.75 & 48.73 & 87.65 & 66.33 & 19 & 71.04  \\
    B\'{e}zierLaneNet\textsuperscript{\S} & ResNet34 &  49.24 & 75.57 & 53.91 & 91.59 & 73.2  & 69.2  & 76.74 & 48.05 & 87.16 & 62.45 & 888 & 69.9  \\
    CLRNet\cite{zheng2022clrnet}\textsuperscript{\dag} & ResNet18 &  55.23 & 79.58 & 62.1  & 93.3  & 78.33 & 73.71 & 79.66 & 53.14 & 90.25 & 71.56 & 1321  & 75.11 \\
    CLRNet\cite{zheng2022clrnet}\textsuperscript{\dag} & ResNet101 &  55.55 & 80.13 & 62.96 & 93.85 & 78.78 & 72.49 & 82.33 & 54.5 & 89.79 & 75.57 & 1262  & 75.51  \\
    CLRNet\cite{zheng2022clrnet}\textsuperscript{\dag} & DLA34 &  55.64 & 80.47 & 62.78 & 93.73 & 79.59 & 75.3 & 82.51 & 54.58 & 90.62 & 74.13 & 1155  & 75.37\\
    \Xhline{0.5px}
    B\'{e}zierFormer & ResNet18 &  55.32 & 79.44 & 62.61 & 93.07 & 77.52 & 74.45 & 75.48 & 52.57 & 89.91 & 71.89 & 1580  & 74.39  \\
    B\'{e}zierFormer & Swin-Tiny &  \textbf{57.07} & \textbf{80.63} & \textbf{63.95} & 93.71 & 78.7 & 74.93 & 81.73 & 55.09 & 90.06 & 73.61 & 1025  & 76.93  \\
    \Xhline{1px}
    \end{tabular}%
    \captionsetup{justification=raggedright,singlelinecheck=false}
    \caption{Results on CULane test split. \textsuperscript{\dag} means the results are referred to \cite{zheng2022clrnet}, \textsuperscript{\ddag} means the results are from the corresponding original papers, and \textsuperscript{\S} means the results are reproduced based on official code.}
\label{tab_culane}%
\end{table*}%

\subsection{Qualitative Results}
\begin{figure*}[h]
\centering
\includegraphics[width=1.0\textwidth]{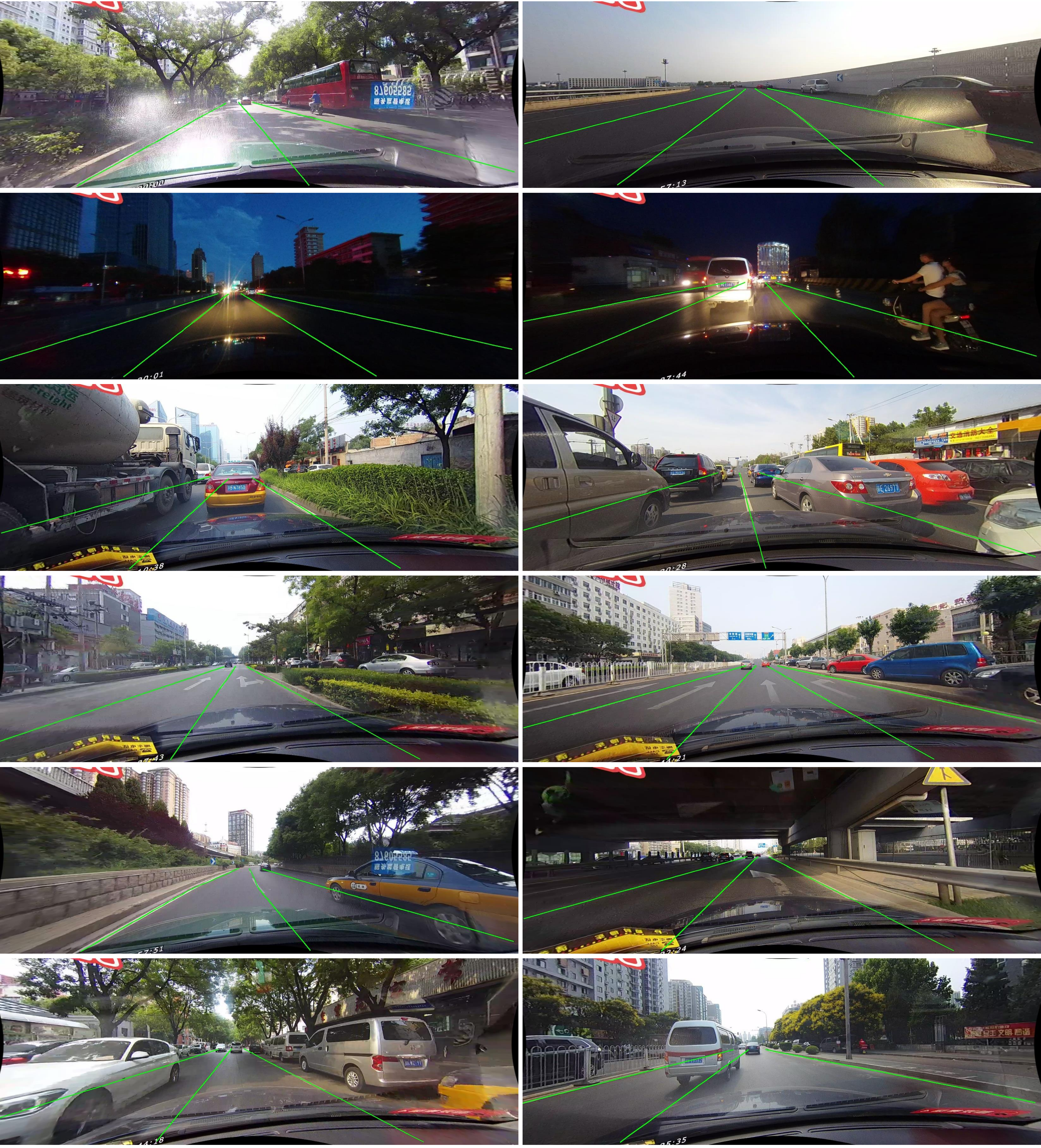} 
\caption{Qualitative results of B\'{e}zierFormer on CULane, including the challenging scenarios like Highlight, Night, Crowd, Arrow, Curve, Shadow and No Line.}
\label{fig_culane}
\end{figure*}

\begin{figure*}[h]
\centering
\includegraphics[width=1.0\textwidth]{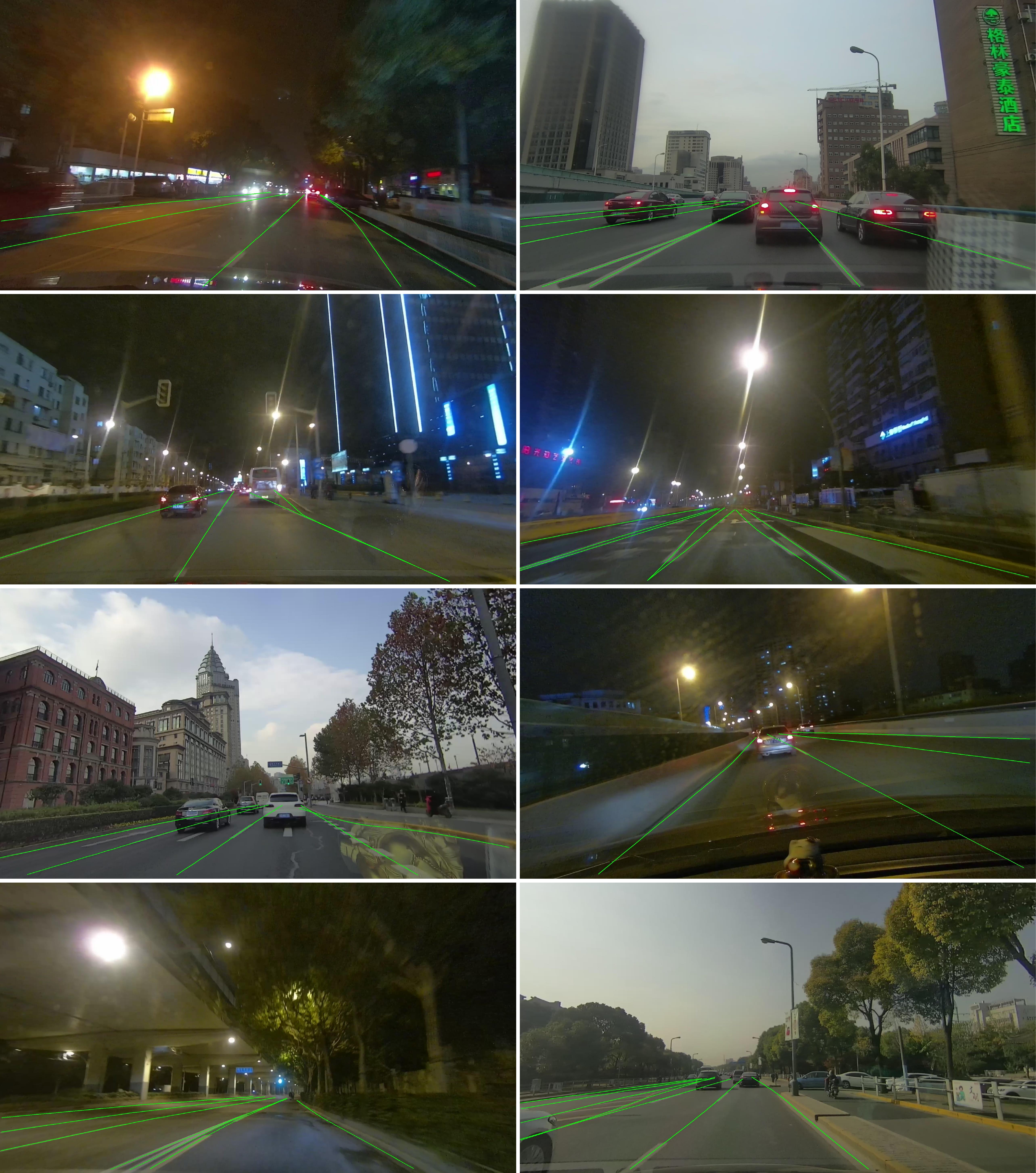} 
\caption{Qualitative results of B\'{e}zierFormer on CurveLanes, including the complex topology like merged lanes, forked lanes, curves lanes, dense lanes and horizontal lanes.}
\label{fig_curvelane}
\end{figure*}

\begin{figure*}[h]
\centering
\includegraphics[width=1.0\textwidth]{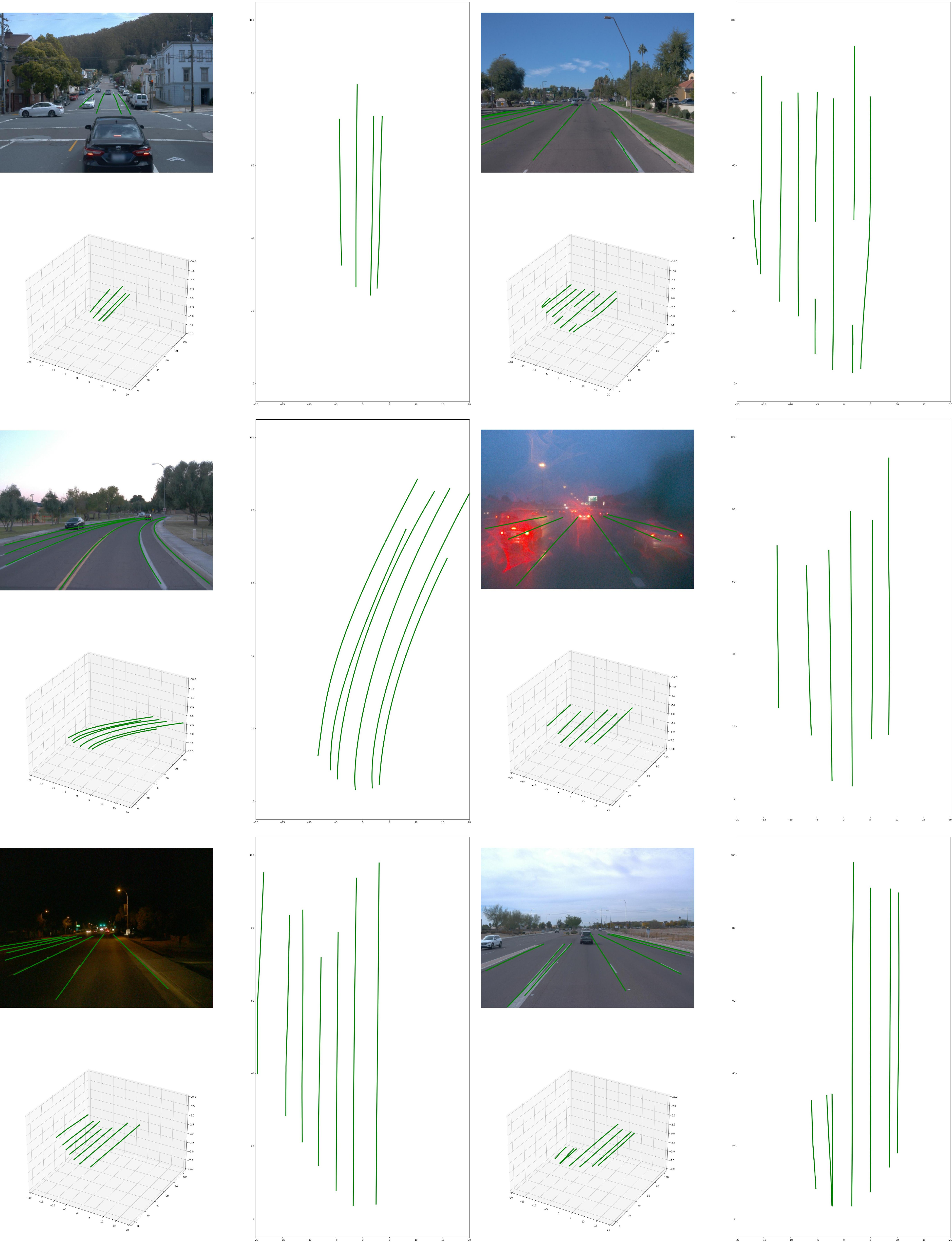} 
\caption{Qualitative results of B\'{e}zierFormer on OpenLane.}
\label{fig_openlane}
\end{figure*}

To have an intuitive understanding of the performance of B\'{e}zierFormer, we give the qualitative results on CULane \cite{pan2018spatial}, CurveLanes \cite{xu2020curvelane} and OpenLane \cite{chen2022persformer}.
As shown in Figure \ref{fig_culane}, B\'{e}zierFormer is robust in challenging scenarios like Highlight, Night, Crowd, Arrow, Curve, Shadow, and No Line.
Figure \ref{fig_curvelane} shows the results on CurveLanes and indicates that  B\'{e}zierFormer performs well in complex topologies like merged lanes, forked lanes, curves lanes, dense lanes, and nearly horizontal lanes.
Figure \ref{fig_openlane} draws the results on OpenLane and proves the effectiveness of B\'{e}zierFormer's 3D lane detection in different scenes.

\bibliographystyle{IEEEbib}
\bibliography{main}

\end{document}